%% file: acl.tex
\pdfoutput=1

\documentclass[11pt]{article}

\usepackage[]{acl}

\usepackage{times}
\usepackage{latexsym}

\usepackage[T1]{fontenc}

\usepackage[utf8]{inputenc}

\usepackage{microtype}

\usepackage{algorithm}
\usepackage{algorithmic}
\usepackage{amsfonts}
\usepackage{graphicx}
\usepackage{enumitem}
\usepackage{diagbox}

\title{Reinforcement Guided Multi-Task Learning Framework for Low-Resource Stereotype Detection}

\author{Rajkumar Pujari$^{1}$ \and Erik Oveson$^{2}$ \and Priyanka Kulkari$^{2}$ \and Elnaz Nouri$^{3}$\\
  $^{1}$Purdue University \Thanks{ This work is a part of summer internship at Microsoft Research, Redmond} \hfill $^{2}$Microsoft, Redmond \hfill $^{3}$Microsoft Research, Redmond\\
  \small{\texttt{rpujari@purdue.edu}} \hfill \small{\texttt{\{erikov,priyak,elnouri\}@microsoft.com}}
}

\begin{document}
\maketitle
\begin{abstract}
As large Pre-trained Language Models (PLMs) trained on large amounts of data in an unsupervised manner become more ubiquitous, identifying various types of bias in the text has come into sharp focus. Existing `\textit{Stereotype Detection}' datasets mainly adopt a diagnostic approach toward large PLMs. \citet{blodgett2021stereotyping} show that there are significant reliability issues with the existing benchmark datasets. Annotating a reliable dataset requires a precise understanding of the subtle nuances of how stereotypes manifest in text. In this paper, we annotate a focused evaluation set for `\textit{Stereotype Detection}'  that addresses those pitfalls by de-constructing various ways in which stereotypes manifest in text. Further, we present a multi-task model that leverages the abundance of data-rich neighboring tasks such as hate speech detection, offensive language detection, misogyny detection, etc.,  to improve the empirical performance on `\textit{Stereotype Detection}'. We then propose a reinforcement-learning agent that guides the multi-task learning model by learning to identify the training examples from the neighboring tasks that help the target task the most. We show that the proposed models achieve significant empirical gains over existing baselines on all the tasks.
\end{abstract}

\section{Introduction}\label{sec:intro}

\input{introduction}

\section{Related Work}\label{sec:reld_work}

\input{related_work}

\section{Our Dataset}\label{sec:dataset}

\input{dataset}

\section{Model}\label{sec:model}
\input{model}

\section{Experiments}\label{sec:expt}

\input{experiments}

\section{Results}\label{sec:res}
\input{results}

\section{Analysis \& Discussions}\label{sec:analysis}

\input{analysis}

\section{Conclusion}\label{sec:conc}

\input{conclusion}

\section*{Acknowledgments}\label{sec:acks}
\input{acknowledgments}

\bibliography{anthology, custom}
\bibliographystyle{acl_natbib}




\end{document}

%% file: introduction.tex
Following \citet{elmo-2018} and then \citet{transformer-2017}, several empirically effective large pre-trained language models (PLMs) \citep{bert-2019,xlnet-2019,roberta-2019,bart2020,gpt3-2020} have been proposed. Their empirical success on several existing tasks has resulted in them being ubiquitously used in many NLP applications that interact with humans on a daily basis \citep{olteanu2020search}. As they are trained in an unsupervised manner on large amounts of arbitrary web data, harmful language and bias creeps into their output. This, in turn, is reflected in the applications that are built on top of them and further propagated in society through the use of those applications \citep{Berk2017AnIA}. This issue motivates the need for a two-pronged solution: 1) to diagnose and de-noise the bias in the PLMs and 2) to identify and regulate harmful text externally at the output level. Progress on the first front is necessary to de-bias the PLMs and prevent subtle biases that creep in through their distributed representations which can often only be recognized in retrospect. This results in deep distrust of such systems among general population. The second front is paramount to detect and potentially block harmful output and also ensure that future iterations and novel proposals of the language models don't take spurious paths to gaming the de-biasing techniques. In this work, we focus on identifying harmful text with stereotypical associations externally.

\noindent\fbox{\parbox{\dimexpr\linewidth-2\fboxsep-2\fboxrule\relax}{
$S_1$: She may or may not be a jew but she's certainly cheap! (\textit{insult, stereotype})\\
$S_2$: Burn in hell, you Asian bastard! (\textit{abuse, stereotype})\\
$S_3$: Asians are good at math. (\textit{stereotype})\\
$S_4$: My African-American friend owns a watermelon patch. (\textit{stereotype})
}}

There exist several types of harmful language such as hate-speech, misogyny, stereotypes, abuse, threats, insult etc,. Each type of offensive language has subtle linguistic nuances that are specific to the type of offensive language. Often, offensive text contains multiple types of offense. From the examples above, consider $S_1$ and $S_2$. Both, consist of multiple modes of offense. While $S_3$ is purely a stereotype, it is still undesirable to be perpetuated.

\citet{cardwell1996dictionary} defines \textit{stereotype} as a ``\textit{fixed, over-generalized belief about a particular group or class of people}''. Stereotypes differ from other types of offensive text in two key aspects: ($1$) they require knowledge of their existence in the society to be identified, and ($2$) they might also often express positive sentiment about the target group. Although some stereotypes ostensibly express positive sentiment towards the target group, they are still undesirable as they propagate false biases in the society and are offensive to the target group. Consider sentences $S_3$ and $S_4$ from above examples. While $S_3$ expresses positive sentiment, it is still false and undesirable. $S_4$ requires knowledge of that particular stereotype's history to understand its offensive nature. Requiring prior knowledge makes annotating data for the task of `\textit{Stereotype Detection}' harder, as annotators are unlikely to be aware of all the stereotypes that exist in the society. \citep{positive-stereotypes-2008}. 

\begin{table*}[!htbp]
    \centering
    \resizebox{!}{35pt}{
        \begin{tabular}{l}
            \hline
            \textbf{Examples}\\\hline
            $1$. Ethiopians like stew (\textit{Explicit Stereotype})\\
            $2$. The lawyer misrepresented the situation and tricked the person (\textit{Implicit Stereotypical Association})\\
            $3$. Jews spend money frivolously (\textit{Anti-Stereotypes})\\
            $4$. There is an Asian family that lives down the street (\textit{Non-Stereotypes})\\\hline
        \end{tabular}
    }
    \caption{Examples of Various Categories of Text with Stereotypical Associations}
    \label{tab:stereotype_examples}
\end{table*}

Two recent works have proposed pioneering diagnostic datasets for measuring stereotypical bias of large PLMs \citep{stereoset2020,crowspairs2020}. But, \citet{blodgett-etal-2021-stereotyping} has demonstrated that these datasets suffer from two major types of issues: ($1$) conceptual: include harmless stereotypes, artificial anti-stereotypes, confusing nationality with ethnicity etc, and ($2$) operational: invalid perturbations, unnatural text, incommensurable target groups etc,. In addition, diagnostic datasets also suffer from lack of sufficient coverage of subtle nuances of manifestations of stereotypes in text. This makes them less suitable for training an effective discriminative classifier. Hence, we undertake a focused annotation effort to create a fine-grained evaluation dataset. We mainly aim to alleviate the \textit{conceptual} issues of anti- vs. non-stereotypes, containing irrelevant stereotypes and \textit{operational} issues of unnatural text, invalid perturbations. We achieve this by a mix of ($1$) selecting more appropriate data candidates and ($2$) devising a focused questionnaire for the annotation task that breaks down different dimensions of the linguistic challenge of `\textit{Stereotype Identification}'. Collecting real-world data from the social forum Reddit for annotation also results in better coverage of subtle manifestations of stereotypes in text. 

Although \textit{stereotypes} differ from other types of \textit{offensive language} in multiple ways, they also overlap to a significant extent. Often, various types of offensive text such as abuse, misogyny and hate speech integrally consists stereotypical associations. Abundance of high-quality annotated datasets are available for these neighboring tasks. We leverage this unique nature of \textit{Stereotype Detection} task to propose a multi-task learning framework for all related tasks. As the overlap between the tasks is only partial, we then propose a reinforcement learning agent that learns to guide the multi-task learning model by selecting meaningful data examples from the neighboring task datasets that help in improving the target task. We show that these two modifications improve the empirical performance on all the tasks significantly. Then, we look more closely at the reinforcement-learning agent's learning process via a suite of ablation studies that throw light on its intricate inner workings. To summarize, our main contributions are:
\begin{enumerate}[noitemsep]
    \item We devise a focused annotation effort for \textit{Stereotype Detection} to construct a fine-grained evaluation set for the task.
    \item We leverage the unique existence of several correlated neighboring tasks to propose a reinforcement-learning guided multitask framework that learns to identify data examples that are beneficial for the target task.
    \item We perform exhaustive empirical evaluation and ablation studies to demonstrate the effectiveness of the framework and showcase intricate details of its learning process.\footnote{Our code and data is available at \url{https://github.com/pujari-rajkumar/rl-guided-multitask-learning}}
\end{enumerate}

%% file: related_work.tex
With the rise of social media and hate speech forums online \citep{phadke2020many,szendro2021suicide} offensive language detection has become more important that ever before. Several recent works focus on characterizing various types of offensive language detection \citep{hate-speech-survey-fortuna-2018,misogyny-detection-survey-2019,abuse-detection-survey-pushkar-2019,parekh2017toxic}. But, works that focus solely on \textit{Stereotype Detection} in English language are scarce. This is partly because stereotypes tend to be subtler offenses in comparison to other types are offensive languages and hence receive less immediate focus, and in part due to the challenge of requiring the knowledge of the stereotype's existence in society to reliably annotate data for the task. We approach this problem by breaking down various aspects of stereotypical text and crowd-sourcing annotations only for aspects that require linguistic understanding rather than world-knowledge.

Few recent works have focused solely on \textit{stereotypes}, some proposing pioneering diagnostic datasets \citep{stereoset2020,crowspairs2020} while others worked on knowledge-based and semi-supervised learning based models \citep{fraser-etal-2021-understanding,badjatiya2019stereotypical} for identifying stereotypical text. Computational model based works either use datasets meant for other tasks such as hate speech detection etc, or focus mainly on the available diagnostic datasets modified for classification task. But, diagnostic datasets suffer from lack of sufficient coverage of naturally occurring text due to their crowd-sourced construction procedure \citep{blodgett-etal-2021-stereotyping}. We address these issues in our work by collecting natural text data from social forum Reddit, by mining specific subreddits that contain mainly subtle stereotypical text.

Multi-task learning \citep{caruana1997multitask}, can be broadly classified into two paradigms \citep{ruder2017overview}: hard parameter sharing \citep{caruana1997multitask} and soft parameter sharing \citep{DBLP:journals/corr/YangH16a,duong-etal-2015-low}. We implement hard-parameter sharing based multi-task model for our experiments. 

Given the low-resource setting on \textit{Stereotype Detection} task, semi-supervised data annotation is one plausible solution for the problem. Several recent works have also been focusing on reinforcement-learning guided semi-supervision \citep{ye-etal-2020-zero, konyushkova2020semi,laskin2020reinforcement}. \citet{ye-etal-2020-zero}, in particular, work with a single-task and unsupervised data to generate automated-annotations for new examples. In contrast, we use the data from neighboring tasks with different labels for multi-task learning and apply an RL agent to select examples for training the neighboring task in such a way that benefits the target task the most.

%% file: dataset.tex
As \citet{blodgett2021stereotyping} demonstrates, existing diagnostic datasets such as \textit{Stereoset} \citep{stereoset2020} and CrowS-Pairs \citep{ crowspairs2020} suffer from several conceptual and operational issues. Additionally diagnostic datasets, by nature, also suffer from lack of coverage of subtle manifestations of stereotypes in text. But, this is mandatory in the dataset used to train an effective discriminatory classifier that can generalize well to identifying stereotypical associations in open domain use cases. This is often the case in the examples encountered in downstream applications that interact with humans on a daily basis.

\citet{stereoset2020} and \citet{crowspairs2020} both follow the following dataset construction procedure: they provide a prompt word or phrase, which is the target group of the stereotype, and ask the crowd workers to construct a stereotypical sentence which includes the provided prompt. Then, they either flip the identity of the target group \citep{crowspairs2020}, the associated characteristic/action of the group or both \citep{stereoset2020} to obtain negative (\textit{anti-stereotypes}) examples.

Although this process results in good coverage of existing stereotypes within society due to the variety of prompts provided, it also results in data examples that either explicitly state the stereotype or are anti-stereotypes that are mostly contrived sentences and are unlikely to occur in regular discourse. An example from \textit{Stereoset} dataset \citep{stereoset2020} is shown below to illustrate this point. $1$ is an explicitly states stereotype where as $2$ and $3$ are unlikely to occur in regular discourse.

\noindent\fbox{\parbox{\dimexpr\linewidth-2\fboxsep-2\fboxrule\relax}{
$1$. He is an Arab. Must be a terrorist. (\textit{stereotype})\\
$2$. He is an Arab. Must be a pacifist. (\textit{anti-stereotype})\\
$3$. He is an Arab. Dog needs walk.  (\textit{unrelated})
}}

We address this issue by additionally annotating text from actual discussions from Reddit to create a more representative dataset for the task. We use the collected examples for a crowd-sourced annotation task to build the dataset. The annotation task is explained in further detail below. We use the prompts from \citet{stereoset2020} and \citet{crowspairs2020} to collect candidates from selected subreddits which are likely to contain a good coverage of stereotypical and non-stereotypical text. We also use the explicit stereotypes from the \textit{Stereoset} and \textit{CrowS Pairs} datasets as candidates. Then, we design a focused annotation task for these candidates to identify various types of Stereotypes.

\subsection{Construction Procedure}
We make the key observation that there are three broad classes of text that contain potentially harmful over-simplified beliefs: $1$) text that is mainly intended to express a stereotype (\textit{explicit stereotype}), $2$) text that is not mainly intended to convey a stereotype but nevertheless propagates a stereotypical association (\textit{implicit stereotypical association}), and $3$) text that expresses an over-simplified belief which is \textit{not} that widely-known, hence is not considered stereotypical (\textit{anti-stereotype}). In addition, there is $4$) text that doesn't contain any over-simplified beliefs about a group of people. We call this type of text as \textit{non-stereotypes}. Examples of different types text are shown in table \ref{tab:stereotype_examples}.

\citet{stereoset2020} and \citet{crowspairs2020} are mainly made up of \textit{explicit-stereotypes} and \textit{anti-stereotypes}. They lack coverage of \textit{implicity stereotypical associations} and \textit{non-stereotypes} due to their construction process and the nature of diagnostic datasets. These are necessary to build a task-representative classification dataset. Hence, in our annotation task we aim to add non-stereotypical examples that contain the same prompts as the ones that are used to create the stereotypical examples. To this end, we collect candidate examples from two subreddit forums \textit{/r/Jokes} and \textit{/r/AskHistorians}. We observe that \textit{/r/Jokes} consists of a high percentage of text with stereotypical associations (both \textit{explicit} and \textit{implicit} stereotypes) and \textit{/r/AskHistorians} tends to consist mainly factual text that is carefully constructed to avoid stereotypical associations. We collect examples that contain the prompts provided in the \citet{stereoset2020} dataset as candidates for annotation. We also use the explicit stereotypes from \textit{Stereoset} and \textit{CrowS-Pairs} datasets for annotation. We perform annotation using \textit{Amazon Mechanical Truk}. For each candidate sentence, we ask the annotators to answer the following questions:
\begin{enumerate}[noitemsep]
    \item Is there an over-simplified belief about a particular type of person ``intentionally'' expressed in the text?
    \item Is there an ``unintentional'', widely-known stereotypical association present in the text?
    \item Does the sentence seem made up (unlikely to occur in regular discourse)?
\end{enumerate}
Each example is annotated by three annotators and we use the majority answer as the gold label. This annotation allows us to separate the text into one of the above $4$ categories. Our dataset consists of $742$ explicit stereotypes, $282$ implicit stereotypes and $1,197$ non-stereotypes. We show the summary statistics of the annotated dataset in table \ref{tab:data_stats}.

\subsection{Ethics Statement}
We conducted a qualification test to select workers based on their performance. The workers were paid a bonus of USD $0.10$ for taking the qualification text. We paid USD $0.25$ for a batch of $10$ examples, each batch taking $45$-$60$ seconds on average. This amounts to USD $15-20$/hour. We displayed a warning on the task that said that the task might contain potentially offensive language. We didn't collect any personal identifying information of the workers other than their worker ID for assigning qualifications. We restricted the workers location to the USA with minimum of $5,000$ approved HITs and $98\%$ HIT approval rate.

\begin{table}[htbp]
    \centering
    \begin{tabular}{lc}
    \hline
        \textbf{Data Type} & \textbf{Size} \\\hline
        \textit{Explicit Stereotypes} & $742$ \\
        \textit{Implicit Stereotypes} & $282$ \\
        \textit{Non-Stereotypes} & $1,197$ \\
        \textit{Total Examples} & $2,221$ \\\hline
    \end{tabular}
    \caption{Summary Statistics of Annotated Dataset}
    \label{tab:data_stats}
\end{table}

%% file: model.tex
\noindent As discussed in section \ref{sec:intro}, high-quality gold data for \textit{Stereotype Detection} is scarce. But, several tasks with correlating objectives have abundance of high-quality annotated datasets. We observe that several tasks under the general umbrella of \textit{Offensive Language Detection} such as \textit{Abuse Detection}, \textit{Hate Speech Detection} \& \textit{Misogyny Detection} often include text with stereotypical associations, as demonstrated in examples $S_1$ and $S_2$ in section \ref{sec:intro}. We call these tasks \textit{neighboring tasks}. We leverage the neighboring task datasets to improve the performance on the low-resource setting of \textit{Stereotype Detection}. First, we propose a multi-task learning model for all the tasks. Then, we make the key observation that \textit{``all examples from the neighboring tasks are \textbf{not} equally useful for the target task''} as the objectives only overlap partially. Further, we propose a reinforcement-learning agent, inspired from \citet{ye-etal-2020-zero}, that learns to select data examples from the neighboring task datasets which are most relevant to the target task's learning objective. We guide the agent via reward assignment based on shared model's performance on the evaluation data of the target task. We experiment both the settings with $4$ popular large PLMs as base classifiers and demonstrate empirical gains using this framework.

In subsection \ref{sec:mtl}, we describe the multi-task learning (MTL) model followed by the Reinforcement Learning guided multi-task learning model (RL-MTL) in subsection \ref{sec:rl_mtl}. Then, in subsection \ref{sec:baselines}, we describe the baseline classifiers we use for our experiments.

\subsection{Multi-Task Learning Model}\label{sec:mtl}

The motivation behind our Multi-Task Learning model is to leverage the transfer learning gains from the neighboring tasks to improve the target task. As the tasks have partially overlapping objectives, solving the selected neighboring tasks effectively requires an understanding of largely similar linguistic characteristics as the target task. Hence, leveraging the intermediate representations of the text from the neighboring task to boost the classifier is expected to benefit the target task. 

Following this motivation, our proposed multi-task model consists of a fixed PLM-based representation layer, followed by shared parameters that are common for all the tasks. Then, we add separate classification heads for each task. We implement hard parameter sharing \citep{caruana1997multitask,ruder2017overview} in our model. The shared parameters compute intermediate representations for the text input. These intermediate representations are shared by all the tasks. Parameters for the shared representation layers are first optimized by training on the neighboring tasks. Then, they are leveraged as a more beneficial parameter initialization for training on the target task data.

The input to the multi-task model is the text of the data example and a task ID. Output of the model is predicted label on the specified task. Each task in the model could either be a single-class classification task or a multi-label classification task. Classification heads for single-class classification tasks have a softmax layer after the final layer. Multi-label tasks have a sigmoid layer for each output neuron in the final layer of the classification heads.

First, we jointly train the model on each of the neighboring tasks in a sequential manner. Then, we train the multi-task model on the target task and evaluate it on the test set of the target task. 

\subsection{Reinforcement Learning Guided MTL}\label{sec:rl_mtl}

\begin{figure}[!htbp]
    \centering
    \includegraphics[width=7cm]{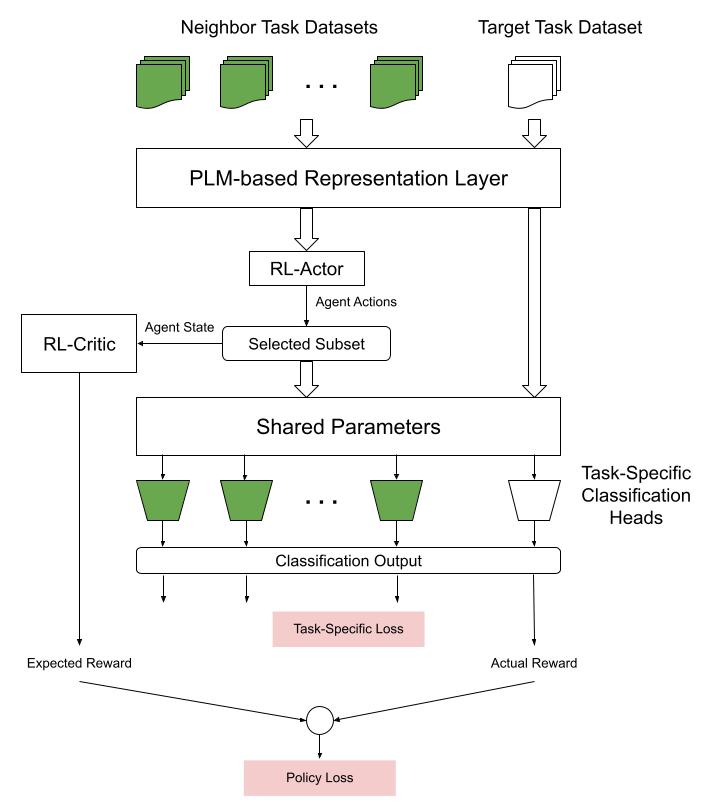}
    \caption{Reinforcement-guided Multi-task Learning Model for Low-Resource Classification Tasks with Correlated Neighboring Tasks}
    \label{fig:rl_mtl_model}
\end{figure}

The RL-guided multi-task model has an additional RL agent on top of the MTL model to select examples from the neighboring task datasets that would be used to train the shared classifier. Key intuition behind the introduction of the RL agent is that, \textit{not all data examples from the neighbor task are equally useful in learning the target task}. Architecture of the RL-guided MTL model is shown in figure \ref{fig:rl_mtl_model}.

Following the above observation, we employ the agent to identify examples that are useful for the target objective and drop examples that distract the classifier from the target task. The agent is trained using an actor-critic reinforcement paradigm \citep{Konda00actor-criticalgorithms}. For each example in the neighbor task, the \textit{Actor} decides whether or not to use it for training the shared classifier. \textit{Critic} computes the expected reward based on \textit{Actor}'s actions for a mini-batch. Upon training using the selected examples, we then assign reward to the agent by evaluating the performance of the shared classifier on the target task. If the $F_1$ scores on the valuation set for $b$ mini-batches, each of size $z$, are \{$F_1^0$, $F_1^1$, $\ldots{}$, $F_1^b$\} and expected rewards predicted by the critic are \{$e_0$, $e_1$, $\ldots{}$, $e_b$\}, then the policy loss is computed as follows:
\begin{equation}
    \hat{F}_1^i = \frac{F_1^i - \mu_{F_1}}{\sigma_{F_1} + \epsilon}
\end{equation}
\begin{equation}
    \textrm{p} = -\frac{1}{b} \Sigma_{i=1}^b (\hat{F}_1^i - e_i) \times \frac{1}{z} \Sigma_{j=1}^{z} log(P[a_j^i]) 
\end{equation}
\begin{equation}
     \textrm{v} = \frac{1}{b} \Sigma_{i=1}^b \textrm{$\mathbb{L}_1$-loss}(1, \hat{F}_1^i)
\end{equation}
\begin{equation}\label{eqn:loss}
    \textrm{total loss} = \textrm{policy loss (p)} + \textrm{value loss (v)}
\end{equation}
where $\epsilon$ is a smoothing constant, $a_j^i$ is the action decided by the Actor for the $j^{th}$ example of mini-batch $i$, $\mu_{F_1}$ and $\sigma_{F_1}$ are mean and standard deviations of the macro-$F_1$ scores, respectively.

The algorithm for RL-guided Multitask learning is shown in algorithm \ref{algo:rl_mtl}. Input to the RL-MTL model is a set of neighboring task datasets and a target task dataset. Output is trained classifier $\mathbb{C}$. We initialize the parameters of the RL-MTL base classifier with the trained parameters of the MTL model. Later, we evaluate the impact of this initialization via an ablation study in section \ref{sec:mtl_impact}.

\begin{algorithm}[htb]
\caption{RL-Guided MTL}
\label{algo:rl_mtl}
\textbf{Require}: Neighbor Datasets \{$\mathbb{N}_{0}$, $\mathbb{N}_{1}$, $\ldots{}$, $\mathbb{N}_{d}$\}, Target Dataset $\mathbb{T}$\\
\textbf{Parameters}: Policy Network $\mathbb{P}$ that includes Actor Network $\mathbb{A}$ and Critic Network $\mathbb{R}$\\
\begin{algorithmic}[1] 
\STATE Select baseline classifier $\mathbb{C}$
\FOR{episode i = $1$, $2$, $\ldots{}$, $e$}
\FOR{neighbor dataset j = $1$, $2$, $\ldots{}$, $d$}
\FOR{mini-batch k =  $1$, $2$, $\ldots{}$, $b$}
\STATE Actor Network $\mathbb{A}$ makes binary SELECT / REJECT decision for each example in $\mathbb{N}_{jk}$
\STATE Critic Network $\mathbb{R}$ computes expected reward based on examples selected by Actor $\mathbb{A}$ = $E[r]^{ijk}$
\STATE Train $\mathbb{C}$ on the SELECTED mini-batch subset $\mathbb{N}^{SEL}_{jk}$
\STATE Evaluate on Target Dataset $\mathbb{T}$ and obtain $F_1$ on target dataset evaluation set $F_1^{ijk}$
\ENDFOR
\STATE Use $F_1^{ijk}$s and $E[r]^{ijk}$s to compute loss according to equation \ref{eqn:loss}
\STATE Update parameters of $\mathbb{A}$ and $\mathbb{R}$
\ENDFOR
\ENDFOR
\STATE \textbf{return} Trained classifier $\mathbb{C}$
\end{algorithmic}
\end{algorithm}

%% file: experiments.tex
We perform experiments on \textit{six} datasets in \textit{three} phases. In the first phase, we experiment with PLM-based fine-tuned classifiers for each task as baselines. In the second phase, we experiment with all the tasks using the  multi-task learning model described in section \ref{sec:mtl}, with each PLM as a base classifier. In the third phase, we train the reinforcement-learning guided multi-task learning framework (section \ref{sec:rl_mtl}) for all the tasks with each of the PLMs as base classifier.

\subsection{Base Classifiers}\label{sec:baselines}

We select four popular PLMs as base classifiers for our empirical experiments, namely, BERT-base, BERT-large \citep{bert-2019}, BART-large \citep{bart2020} and XLNet-large \citep{xlnet-2019}. We use the implementations from \citet{wolf-etal-2020-transformers}'s huggingface transformers library\footnote{\url{https://github.com/huggingface/transformers}} for experimentation. We fine-tune a classification layer on top of representations from each of the PLMs as baseline to evaluate our framework.

\subsection{Datasets} \label{sec:neighbor_data}
We use \textit{six} datasets for our empirical evaluation, namely, Jigsaw Toxicity Dataset, Hate Speech Detection \citep{gibert2018hate}, Misogyny Detection \citep{fersini2018overview}, Offensive Language Detection \citep{hateoffensive}, coarse-grained Stereotype Detection (combination of \textit{Stereoset}, \textit{CrowS-Pairs} and Reddit Data) and finally fine-grained Stereotype Detection Data (as described in section \ref{sec:dataset}). We describe each dataset briefly below.

\noindent \textbf{Hate Speech Detection} \citep{gibert2018hate} dataset consists of $10,944$ data examples of text extracted from Stromfront, a white-supremacist forum. Each piece of text is labeled as either \textit{hate speech} or \textit{not}.

\noindent \textbf{Misogyny Detection} \citep{fersini2018overview} dataset consists of $3,251$ data examples of text labeled with the binary label of being \textit{misogynous} or \textit{not}.

\noindent \textbf{Offensive Language Detection} \citep{hateoffensive} dataset was built using crowd-sourced hate lexicon to collect tweets, followed by manual annotation of each example as one of \textit{hate-speech}, \textit{only offensive language} or \textit{neither}. This dataset contains $24,783$ examples. 

\noindent \textbf{Coarse-Grained Stereotype Detection}: We create this dataset by combining stereotypical examples from \textit{Stereoset} and \textit{CrowS-Pairs} datasets to get positive examples, followed by adding negative examples from the subreddit \textit{/r/AskHistorians}. We do not use crowd sourced labels in this dataset. We use the labels from the original datasets. The dataset consists of $23,900$ data examples.

\noindent \textbf{Fine-Grained Stereotype Detection}: This dataset is the result of our annotation efforts in section \ref{sec:dataset}. It consists of $2,221$ examples, each annotated with one of three possible labels: \textit{explicit stereotype, implicit stereotype and non-stereotype}.

\noindent \textbf{Jigsaw Toxicity Dataset}\footnote{\url{https://tinyurl.com/2vjmprnh}} consists of $159,571$ training examples and $153,164$ test examples labeled with one or more of the \textit{seven} labels: \textit{toxic, severely toxic, obscene, threat, insult, identity hate, none}. We use this data only for training. We don't evaluate performance on this dataset.

%% file: results.tex
\begin{table*}[!htb]
\centering
    \resizebox{!}{!}{
        \begin{tabular}{lccccc}
            \hline
            \textbf{\begin{tabular}[c]{@{}c@{}}Model\end{tabular}} &
            \textbf{\begin{tabular}[c]{@{}c@{}}Hate Speech\\ Detection\end{tabular}} & \textbf{\begin{tabular}[c]{@{}c@{}}Offense\\ Detection\end{tabular}} & \textbf{\begin{tabular}[c]{@{}c@{}}Misogyny\\ Detection\end{tabular}} & \textbf{\begin{tabular}[c]{@{}c@{}}Coarse\\ Stereotypes\end{tabular}} &
            \textbf{\begin{tabular}[c]{@{}c@{}}Fine\\ Stereotypes\end{tabular}}\\ \hline
            BERT-base & $66.47$ & $66.13$ & $74.16$ & $65.71$ & $61.36$ \\
            BERT-large & $67.05$ & $63.90$ & $72.13$ & $59.63$ & $55.42$ \\
            BART-large & $68.91$ & $65.86$ & $73.12$ & $63.40$ & $54.64$ \\
            XlNet-large & $59.14$ & $48.33$ & $63.16$ & $63.71$ & $53.80$ \\\hline
            \textbf{Multi-Task Learning}  &  &  &  &  \\\hline
            BERT-base + MTL & $69.21^{\dagger}$ & $68.57^{\dagger}$ & $73.48$ & $68.29^{\dagger}$ & $65.00^{\dagger}$ \\
            BERT-large + MTL & $69.78^{\dagger}$ & $65.14^{\dagger}$ & $73.94^{\dagger}$ & $61.96^{\dagger}$ & $61.65^{\dagger}$ \\
            BART-large + MTL & $67.79$ & $68.03^{\dagger}$ & $74.40^{\dagger}$ & $65.77^{\dagger}$ & $64.90^{\dagger}$ \\
            XlNet-large + MTL & $61.68^{\dagger}$ & $46.35$ & $64.42^{\dagger}$ & $65.21^{\dagger}$ & $57.00^{\dagger}$ \\\hline
            \textbf{RL-guided MTL}  &  &  &  & \\\hline
            BERT-base + RL-MTL & $72.06^{\dagger}$ & $68.97$ & $74.78^{\dagger}$ & $74.18^{\dagger}$ & $65.72^{\dagger}$ \\
            BERT-large + RL-MTL & $69.82$ & $65.97^{\dagger}$ & $75.21^{\dagger}$ & $70.88^{\dagger}$ & $64.74^{\dagger}$ \\
            BART-large + RL-MTL & $69.60^{\dagger}$ & $66.76$ & $75.14^{\dagger}$ & $74.11^{\dagger}$ & $67.94^{\dagger}$ \\
            XlNet-large + RL-MTL & $61.97$ & $47.60^{\dagger}$ & $63.21$ & $67.98^{\dagger}$ & $56.37$ \\\hline
        \end{tabular}
    }
    \caption{Results on all the Datasets for various phases. Macro-F1 score has been reported. $^{\dagger}$ indicates that improvements over the corresponding model in the previous section are statistically significant according to McNemar's statistical significance test.}
    \label{tab:all_results}
\end{table*}

We present the results of the empirical evaluation tasks in table \ref{tab:all_results}. In \textit{Hate Speech Detection} task, we observe that RL-MTL learning results in significant improvements over all the baseline classifiers. Plain MTL model also improves upon the baseline classifiers except in the case on BART-large. The best model for this task is BERT-base + RL-MTL which achieves a macro-F1 score of $72.06$ compared to $68.91$ obtained by the best baseline classifier. Best MTL model obtains $69.78$ F1.

For \textit{Hate Speech and Offensive Language Detection} task, the respective numbers for baseline, MTL and RL-MTL models are $66.13$, $68.57$ and $68.97$. The models achieve $74.16$, $74.40$ and $75.21$ on \textit{Misogyny Detection} task, respectively. In \textit{Coarse-Grained Stereotype Detection} task, they achieve $65.71$, $68.29$ \& $74.18$, which is a significant gradation over each previous class of models. On our focus evaluation set of \textit{Fine-Grained Stereotype Detection}, we achieve $61.36$, $65.00$ \& $67.94$ in each class of models. The results on this dataset are obtained in a zero-shot setting as we only use this dataset for evaluation.


%% file: analysis.tex
In the first ablation study described in subsection \ref{sec:mtl_impact}, we study the importance of initializing RL-MTL model with the trained parameters of MTL model. Following that, we look into more detail about the usefulness of neighbor tasks on the target task via an ablation study.
We describe these experiments in further detail in subsection \ref{sec:task_ablation}.

\subsection{Impact of MTL Prior on RL-MTL} \label{sec:mtl_impact}

In our original experiments, we initialize the parameters of RL-MTL model with trained parameters from the MTL model. This allows the RL agent to begin from a well-optimized point in the parameter sample space. In this ablation study, we initialize the RL-MTL model from scratch to see how it impacts the performance of the RL-MTL model. We perform this experiment with BERT-base as base classifier. The performance of the RL- MTL model without initialization drops to $70.23$ on HS task, $67.23$ on HSO task, $71.10$ on MG task, $60.42$ on CG-ST task and $57.32$ on FG-ST task. The respective numbers for the MTL initialized model are $72.06$, $68.97$, $74.78$, $74.18$ and $65.72$. Initialization has biggest impact on the \textit{Coarse-} and \textit{Fine-Grained Stereotype Detection} tasks. Overall, initialization with MTL trained parameters results in a better convergence point for the RL-MTL model.

\subsection{Neighbor-Task Ablation Study}\label{sec:task_ablation}
In this task, we aim to study the neighbor tasks that are most useful for each target task. For each dataset, we train RL-MTL framework with only one other neighbor dataset. We see which task yields biggest improvement for each target task. We experiment with various combinations of datasets for this dataset. Results for this ablation study are shown in table \ref{tab:task_ablation}. All experiments in this ablation study are performed using BERT-base as the base classifier.

\begin{table}[htbp]
    \centering
    \begin{tabular}{|l|c|c|c|c|}
        \hline
        \backslashbox{\textbf{T}}{\textbf{N}} & \textbf{HS} & \textbf{HSO} & \textbf{MG} & \textbf{C-ST} \\\hline
        \textbf{HS} & - & $69.69$ & $70.07$ & \textbf{$71.10$} \\\hline
        \textbf{HSO} & $66.71$ & - & $66.56$ & \textbf{$67.39$} \\\hline
        \textbf{MG} & $70.98$ & \textbf{$75.87$} & - & $73.89$ \\\hline
        \textbf{C-ST} & $66.15$ & \textbf{$67.40$} & $63.82$ & - \\\hline
        \textbf{F-ST} & $63.80$ & \textbf{$63.65$} & $59.94$ & $56.12$ \\\hline
    \end{tabular}
    \caption{Macro-F1 scores on each Target Task in Task Ablation Study for each individual Neighbor Task. T: \textit{Target Task}, N: \textit{Neighboring Task}, HS: \textit{Hate Speech Detection}, HSO: \textit{Hate Speech and Offensive Language Detection}, MG: Misogyny Detection, C-ST: \textit{Coarse-Grained Stereotype Detection}, F-ST: \textit{Fine-Grained Stereotype Detection}}
    \label{tab:task_ablation}
\end{table}

Results in table \ref{tab:task_ablation} show that for both \textit{Hate Speech Detection} (HS) and \textit{Hate Speech and Offensive Language Detection} (HSO) tasks, \textit{Coarse-Grained Stereotype Detection} (C-ST) neighboring task yields the best improvements to $71.1$ and $67.39$ macro-F1, respectively. All the other three neighboring tasks are useful in improving the performance of the base classifier from $66.47$ and $66.13$ F1 scored. For \textit{Misogyny Detection} (MG) task, HSO neighboring task results in an improvement from $74.16$ to $75.87$, while the other two tasks deteriorate the performance on the task. It is also interesting to note that, the combined performance on the task with all three datasets is lower ($74.78$) than when using HSO data alone. For both \textit{Coarse-} and \textit{Fine-grained Stereotype Detection} (F-ST) tasks, HS and HSO datasets improve the performance over the baseline, while MG deteriorates the performance. The combined improvement of all the neighboring tasks together is higher than either HS or HSO neighboring tasks alone. It is also interesting to note that the C-ST task doesn't contribute significantly to performance improvement on F-ST task. This might be due to the presence of anti-stereotypes and several other issues pointed out in \citet{blodgett-etal-2021-stereotyping}.



%% file: conclusion.tex
We tackle the problem of \textit{Stereotype Detection} from \textit{data annotation} and \textit{low-resource computational framework} perspectives in this paper. First, we discuss the key challenges that make the task unique and a low-resource one. Then, we devise a focused annotation task in conjunction with selected data candidate collection to create a fine-grained evaluation set for the task. 

Further, we utilize several neighboring tasks that are correlated with our target task of \textit{'Stereotype Detection'}, with an abundance of high-quality gold data. We propose a reinforcement learning-guided multitask learning framework that learns to select relevant examples from the neighboring tasks that improve performance on the target task. Finally, we perform exhaustive empirical experiments to showcase the effectiveness of the framework and delve into various details of the learning process via several ablation studies.

%% file: acknowledgments.tex
We thank the anonymous reviewers and meta-reviewer for their insightful comments that helped in improving our paper.